\documentclass[letterpaper]{article} 
\usepackage{aaai2026}  
\usepackage{times}  
\usepackage{helvet}  
\usepackage{courier}  
\usepackage[hyphens]{url}  
\usepackage{graphicx} 
\def\UrlFont{\rm}  
\usepackage{natbib}  
\usepackage{caption} 
\usepackage{algorithm}
\usepackage{algorithmic}
\usepackage{amsmath}
\usepackage{amsmath}
\usepackage{amssymb}
\usepackage[switch]{lineno}
\usepackage{booktabs}
\usepackage{multirow}
\usepackage{newfloat}
\usepackage{listings}
\DeclareCaptionStyle{ruled}{labelfont=normalfont,labelsep=colon,strut=off} 
\floatstyle{ruled}
\newfloat{listing}{tb}{lst}{}
\floatname{listing}{Listing}
\title{From Intent to Execution: Multimodal Chain-of-Thought Reinforcement Learning for Precise CAD Code Generation
}

\author{
    Ke Niu\textsuperscript{1}\dag, 
    Haiyang Yu\textsuperscript{1$\ast$}\dag, 
    Zhuofan Chen\textsuperscript{1}\dag,
    Mengyang Zhao\textsuperscript{1}, 
    Teng Fu\textsuperscript{1,2}, 
    Bin Li\textsuperscript{1}\thanks{Corresponding author, \dag Equal contribution.},\\
    Xiangyang Xue\textsuperscript{1} \\
    \textsuperscript{1}Fudan University, Shanghai, China. 
    \textsuperscript{2}ByteDance Inc.
    \\
{\tt\small \{kniu22, zfchen23, tfu23\}@m.fudan.edu.cn}\\
{\tt\small \{hyyu20, myzhao20, libin, xyxue\}@fudan.edu.cn}}

\usepackage{bibentry}

\begin{document}

\maketitle

\begin{abstract}
Computer-Aided Design (CAD) plays a vital role in engineering and manufacturing, yet current CAD workflows require extensive domain expertise and manual modeling effort. Recent advances in large language models (LLMs) have made it possible to generate code from natural language, opening new opportunities for automating parametric 3D modeling. However, directly translating human design intent into executable CAD code remains highly challenging, due to the need for logical reasoning, syntactic correctness, and numerical precision.
In this work, we propose CAD-RL, a multimodal Chain-of-Thought (CoT) guided reinforcement learning post training framework for CAD modeling code generation. Our method combines CoT-based Cold Start with goal-driven reinforcement learning post training using three task-specific rewards: executability reward, geometric accuracy reward, and external evaluation reward. To ensure stable policy learning under sparse and high-variance reward conditions, we introduce three targeted optimization strategies: Trust Region Stretch for improved exploration, Precision Token Loss for enhanced dimensions parameter accuracy, and Overlong Filtering to reduce noisy supervision.
To support training and benchmarking, we release ExeCAD, a noval dataset comprising 16,540 real-world CAD examples with paired natural language and structured design language descriptions, executable CADQuery scripts, and rendered 3D models. Experiments demonstrate that CAD-RL achieves significant improvements in reasoning quality, output precision, and code executability over existing VLMs.
\end{abstract}

\section{Introduction}

\begin{figure*}[t]
    \centering
    \includegraphics[width=1\linewidth]{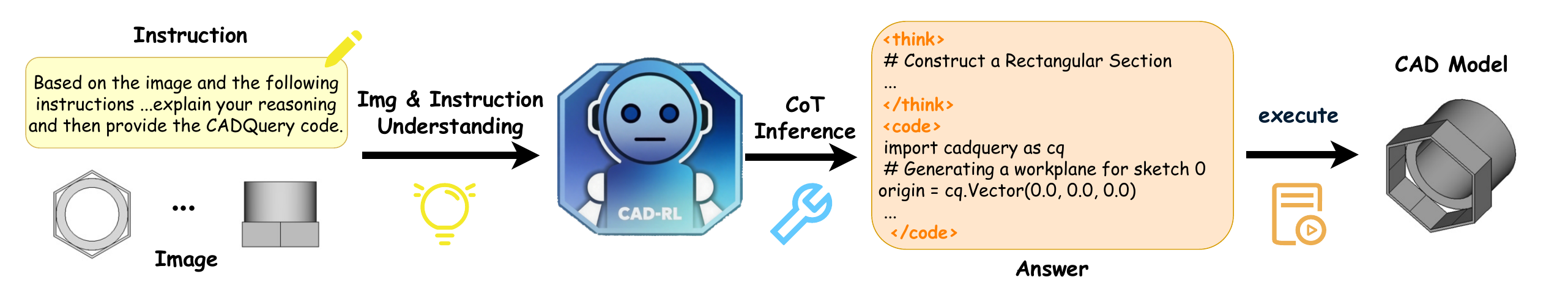}
\caption{Workflow of real-world CAD modeling enabled by CAD-RL.}
\label{1}
\end{figure*}

Computer-Aided Design (CAD) systems are foundational tools in engineering and manufacturing, enabling the creation of precise, parameterized 3D models for a wide range of industrial applications. However, traditional CAD workflows are highly manual, requiring extensive domain expertise and design iteration effort. While prior work has attempted to automate 3D reconstruction from sketches or images~\cite{jayaraman2022solidgen,li2020sketch2cad,wu2021deepcad}, the resulting models are typically non-editable, approximate, and lack parametric control—limitations that are incompatible with the tight tolerance requirements of industrial design.
Recent advances in vision-language models (VLMs)~\cite{niu2025creft,niu2025pht}, particularly in natural language to code generation, open up a new paradigm: transforming high-level design intent into executable and semantically grounded CAD code. This capability promises to enable customized manufacturing, interpretable geometry modeling, and a substantial reduction in CAD development cost, while also bridging the gap between natural communication and formal design languages.

Despite this promising direction, translating complex human intent into executable CAD modeling code remains highly challenging. First, it necessitates understanding precise dimensions and physical constraints embedded in design language. Second, the model must generate code that is syntactically valid, structurally coherent, and functionally executable.
Recent efforts begin to tackle this space. Text-Guided CAD-Coder\cite{he2025cad} maps language prompts to pre-defined command sequences, but its rigid structure restricts generalization to a narrow set of shapes. VLM CAD-Coder\cite{doris2025cad} utilizes single-modality image input to generate CAD code, but the lack of accurate dimensional constraints leads to coarse approximations that fall short in real industrial practice. CAD-Coder~\cite{guan2025cad} achieves high accuracy by generating structured CadQuery code from expert-written descriptions, but its reliance on expert-level inputs hinders accessibility for general users. Furthermore, all of these methods operate on single-modal input, limiting adaptability to diverse, real-world design scenarios.

Existing approaches for adapting VLMs to domain-specific tasks rely primarily on two strategies: supervised fine-tuning (SFT) and reinforcement learning (RL). SFT provides stable, token-level supervision, allowing models to learn syntactic and structural patterns from annotated data. However, it falls short in optimizing downstream task performance—such as geometric accuracy and code executability—and requires large amounts of high-quality annotated data, which is prohibitively expensive in the CAD domain. In contrast, RL enables task-specific, goal-directed optimization using custom reward functions, making it a powerful complement to SFT.
In this work, we posit three key principles for CAD code generation from human intent:
(1) CAD modeling requires strong logical reasoning ability to align abstract design goals with precise, executable operations;
(2) CAD code combines rigid structural syntax (e.g., extrusion) with fine-grained parameters (e.g., dimensions, tolerances), necessitating both syntactic correctness and numerical precision
(3) the same geometry can be realized via multiple code sequences, so the model should prefer concise and efficient representations over verbose or redundant ones.

\begin{figure*}[t]
    \centering
    \includegraphics[width=1\linewidth]{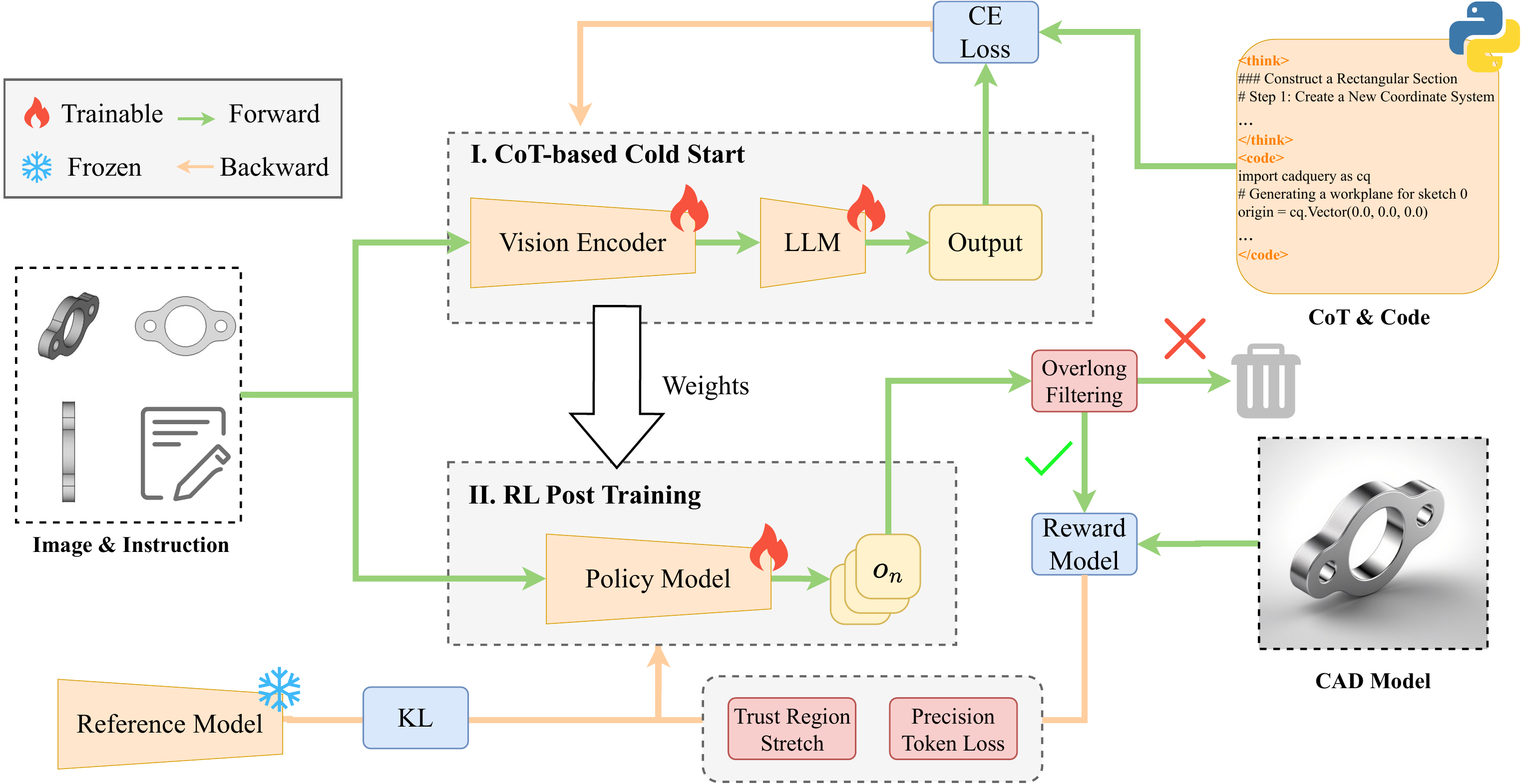}
\caption{Overall architecture of our proposed CAD-RL framework. }
\label{2}
\end{figure*}

To address these challenges, we propose CAD-RL, a multimodal Chain-of-Thought (CoT)-driven reinforcement learning post training framework for CAD code generation. Fig.~\ref{1} illustrates the real-world CAD modeling workflow enabled by CAD-RL, where users provide intent and the system generates precise, executable CadQuery code that compiles into accurate 3D CAD models.
Our method enables robust interpretation of diverse textual design inputs—including natural language and structured specifications—while accurately capturing geometry, topology, and parametric logic. It generates structurally consistent, executable CadQuery code that compiles into high-fidelity 3D CAD models.
Crucially, CAD-RL supports flexible multimodal input interfaces. It can process both natural language descriptions accessible to non-experts and technical specifications preferred by CAD professionals. In addition, our model accepts reference images to further enhance modeling fidelity. This flexibility enables the system to accommodate a wide range of real-world usage scenarios.
As shown in Fig.~\ref{2}, our training pipeline comprises two stages:
(1) CoT-based Cold Start: We curate a small but high-quality dataset of multimodal examples with annotated reasoning traces to help the model learn the structure and format of long-horizon CoT reasoning.
(2) Reinforcement Learning Post Training: We introduce three targeted rewards to guide the exploration of diverse reasoning paths. 
To further support the reasoning quality and output robustness, we propose three fine-grained optimization techniques:
Trust Region Stretch, which expands the policy update range to encourage broader exploration of reasoning trajectories;
Precision Token Loss, which increases gradient weight on numerically sensitive tokens (e.g., dimensions, constraints);
Overlong Filtering, which truncates overly verbose outputs to mitigate noisy reward gradients.
Another key bottleneck for progress in CAD code generation is the lack of high-quality, open-source datasets that reflect industrial-level requirements. To address this, we introduce ExeCAD, a noval benchmark tailored for executable and editable CAD code generation. The dataset includes 16,540 instances, each containing:
(1) a simple natural language prompt simulating non-expert user intent;
(2) a detailed expert-level specification reflecting precise design constraints;
(3) corresponding executable CADQuery code; and
(4) a rendered 3D CAD model.
ExeCAD enables evaluation under varying input modalities and user expertise levels, and provides a realistic foundation for training and benchmarking future text-to-CAD generation systems.

 Our main contributions are:
\begin{itemize}
\item We propose CAD-RL, a novel multimodal CoT-guided RL post training framework for CAD modeling code generation.
\item We introduce three task-specific training strategies to enhance reasoning precision and policy robustness.
\item We release ExeCAD, a high-quality, multi-perspective dataset for executable CAD code generation, supporting non-expert and expert-level input settings.
\end{itemize}

\section{Related Work}
\subsection{RL Adaptation for Code-Oriented VLMs}
Code generation from natural language is a central capability of modern LLMs~\cite{jiang2024survey}. While supervised fine-tuning (SFT) based on maximum likelihood helps models learn syntax and structure, it does not directly optimize for downstream metrics such as execution correctness or semantic alignment.
To address this, reinforcement learning (RL) is increasingly employed to align generation with task-specific objectives~\cite{fujimoto2019off}. RL-based techniques for code LLMs generally fall into three categories:
Policy gradient methods like CodeRL~\cite{le2022coderl} and LEDEX~\cite{jiang2024ledex} use PPO~\cite{schulman2017proximal} to maximize execution-based rewards, but require extensive computation and careful reward design.
Preference-based methods reframe RLHF as a binary classification task over response pairs. DPO~\cite{rafailov2023direct} improves training efficiency via closed-form policy updates but struggles with nuanced preference modeling.
Gradient-based reward optimization, such as DeepSeek-RL~\cite{guo2024deepseek}, scales to large models using in-batch advantage estimation. Hybrid approaches like Swe-RL~\cite{wei2025swe} and StepCoder~\cite{dou2024stepcoder} combine supervised and reward-driven objectives to improve reasoning and correctness.
Each method offers trade-offs between supervision cost, scalability, and alignment fidelity, making them complementary in different stages of model adaptation.

\subsection{CAD Code Generation}
CAD code generation aims to create precise, editable 3D models directly from user instructions. This approach improves on conventional 3D reconstruction by enabling parametric control and design flexibility.
Several methods tackle this task using different modalities and representations. CAD-MLLM~\cite{xu2024cad} and GenCAD~\cite{alam2024gencad} map image inputs to CAD commands using vision-language models. Img2CAD~\cite{you2024img2cad} separates structure prediction and parameter regression in a two-stage design. CAD2Program~\cite{wang20252d} introduces a more flexible code format for richer modeling expressiveness.
Recent methods increasingly adopt CadQuery, a Python-based modeling language, as the output format~\cite{he2025cad,doris2025cad,guan2025cad}. Unlike command sequences or DSLs, CadQuery supports loops, conditionals, and parameterization, making it both expressive and executable. Its compatibility with industrial formats (e.g., STEP, STL) makes it a practical bridge between intent and deployable design.

\section{Method}

\subsection{Overview}

Inspired by recent advancements in RL for code generation~\cite{guo2024deepseek,dou2024stepcoder}, we propose CAD-RL, a reinforcement learning post training framework tailored for executable, high-precision CAD code generation. CAD-RL takes flexible multimodal inputs, including both natural language descriptions (e.g., from non-expert users) and structured design language (e.g., from professional engineers), optionally augmented by reference images. This flexible input paradigm makes our system adaptable to diverse, real-world CAD design scenarios.

The model directly produces executable Python-based CadQuery scripts, enabling parametric modeling with high editability, syntactic correctness, and geometric fidelity. A key innovation in CAD-RL is the integration of Chain-of-Thought (CoT) reasoning into the RL post trainingg loop, aiming to activate pretrained vision-language models’ latent capacity for long-horizon symbolic reasoning.

Our training pipeline (Fig.~\ref{2}) consists of two stages:
\begin{itemize}
    \item \textbf{CoT-based Cold Start} (Sec.~\ref{3.2}): We fine-tune the model on a curated multimodal dataset with expert CoT annotations, teaching the model to map abstract geometric intent to executable CAD programs.
    \item \textbf{Reinforcement Learning Post Training} (Sec.~\ref{3.3}): We optimize task-specific reward signals, executability, geometric accuracy, and external evaluation, to drive exploration beyond imitation and elicit semantically grounded reasoning paths.
\end{itemize}

To further stabilize learning under sparse feedback, we introduce three fine-grained optimization strategies: (1) Trust Region Stretch for expanding exploration bounds; (2) Precision Token Loss for emphasizing numerically sensitive tokens; and (3) Overlong Filteringto reduce reward noise from truncated sequences.

Lastly, we introduce ExeCAD (Sec.~\ref{3.4}), a noval benchmark with 16,540 aligned instances featuring paired natural and structured prompts, executable CAD code, and 3D renderings—providing strong supervision for real-world CAD generation.

\subsection{CoT-based Cold Start}
\label{3.2}

While vision-language models (VLMs) exhibit latent capacity for symbolic reasoning, their pretraining lacks supervision on task-specific, executable representations like CAD code. Consequently, their outputs often lack structural consistency or precise semantics.

To address this, we introduce a supervised CoT-based Cold Start stage that fine-tunes the model on a small, high-quality dataset annotated with expert Chain-of-Thought traces. These multimodal examples guide the model to structure long-horizon reasoning steps that bridge abstract intent and concrete code.

We model the joint reasoning and code sequence using an autoregressive objective:
\begin{equation}
\mathcal{L}_{\text{Cold Start}}(\theta) = - \mathbb{E}_{(x, y, r, c) \sim \mathcal{D}} \left[ \sum_{t=1}^{|s|} \log p_\theta(s_t \mid s_{<t}, x, y) \right]
\end{equation}
where \( \theta \) denotes model parameters, and \( s \) includes both reasoning tokens and CadQuery code. Special delimiter tokens (e.g., \texttt{<Think>}) are inserted to encourage a two-part generation pattern.

\subsection{Reinforcement Learning Post Training}
\label{3.3}

To further improve semantic alignment and execution fidelity, we introduce a reinforcement learning (RL) post training stage that shifts the model beyond imitation learning. In this stage, the model is viewed as a policy \( \pi_\theta \), generating token sequences \( s = [s_1, \dots, s_T] \) conditioned on multimodal input \( (x, y) \), where each output contains both CoT reasoning and CAD code. We optimize the expected reward rather than token likelihood:

\begin{equation}
\mathcal{L}_{\text{RL}}(\theta) = - \mathbb{E}_{s \sim \pi_\theta(\cdot \mid x, y)} \left[ R(s) \right]
\end{equation}

 The reward function \( R(s) \) is composed of three complementary components:
 
 \textbf{Executability Reward} \( R_{\text{exec}} \): This component evaluates whether the generated CAD code is syntactically valid and functionally executable within a standard Python runtime environment. Given that the final output of the model is CadQuery-based Python code, we employ Python interpreter to check for syntax and runtime exceptions.
If the code parses correctly and completes execution without exceptions—a reward of 1 is assigned; otherwise, the reward is 0. This binary structure enforces a hard validity constraint on model outputs, filtering out structurally broken or incomplete programs.
Importantly, we design \( R_{\text{exec}} \) as a multiplicative gating factor for the overall reward function. It ensures that if a program is not executable, all geometric and external evaluation rewards are masked, thereby preventing reward leakage from malformed samples.
Formally, we define executability as:
\begin{equation}
R_{\text{exec}}(c) = 
\begin{cases}
1, & \text{if } \texttt{Execute}(c) \text{ succeeds without error} \\
0, & \text{otherwise}
\end{cases}
\end{equation}

 \textbf{Geometric Accuracy Reward} \( R_{\text{geom}} \): 
This reward evaluates the geometric fidelity of the generated 3D CAD model with respect to the reference design. We execute the generated code to output 3D model and compare it against the ground-truth CAD model using Intersection-over-Union (IoU), a strict volumetric similarity metric defined as the ratio of the intersected volume to the union of both volumes.
Formally, the reward is computed as:
\begin{equation}
R_{\text{geom}}(M_{\text{gen}}, M_{\text{gt}}) = \frac{|M_{\text{gen}} \cap M_{\text{gt}}|}{|M_{\text{gen}} \cup M_{\text{gt}}|}
\end{equation}

\textbf{External Evaluation Reward} \( R_{\text{eval}} \): 
To complement the syntactic and geometric criteria captured by \( R_{\text{exec}} \) and \( R_{\text{geom}} \), we introduce an auxiliary reward that reflects high-level functional correctness and semantic fidelity from a task-centric perspective. This reward is derived from an external evaluator—specifically, GPT-4o~\cite{achiam2023gpt}—which provides a soft, interpretable score assessing how well the generated CAD code aligns with the design intent.

To ensure that the model learns not only syntactically valid or geometrically similar outputs but also semantically faithful and logically coherent CAD programs, we incorporate targeted failure-mode analysis. Based on empirical observations from the CoT-based cold-start stage, we impose stronger penalties on two particularly harmful error categories:
{Reference Frame Misalignment and {Parametric Misassignment. During evaluation, GPT-4o is explicitly prompted to detect such violations and apply severe deductions to the score. This structured evaluation encourages the model to internalize and prefer functionally correct generation patterns, especially in scenarios with underspecified or ambiguous natural language prompts, thus reinforcing alignment between abstract design intent and concrete CAD behavior. Formally, the reward is computed as:
\begin{equation}
R_{\text{eval}} = \text{Norm}({\texttt{Score}(c_{\text{gen}}, x)})
\end{equation}

\noindent
The total reward is defined as:
\begin{equation}
R = R_{\text{exec}}(c) \cdot \left[ \lambda_{\text{geom}} R_{\text{geom}} + \lambda_{\text{eval}} R_{\text{eval}} \right]
\end{equation}

\textbf{Optimization Techniques.} To ensure stable policy learning under sparse and high-variance reward conditions, we introduce three targeted optimization strategies:

\paragraph{(1) Trust Region Stretch.}  
Standard policy gradient methods, such as PPO~\cite{schulman2017proximal}, constrain policy updates using a trust region to ensure training stability. This constraint prevents excessive divergence between the current policy \( \pi_\theta \) and a reference policy \( \pi_{\text{ref}} \), often initialized from a pretrained or SFT model. 

However, in the context of CAD modeling, where code generation involves combinatorial compositions of geometric primitives and operations, such constraints may prematurely collapse exploration. As training progresses, the model tends to converge to narrow, deterministic policies that repeatedly sample nearly identical responses. This phenomenon is particularly detrimental to CoT-based reasoning, where multiple valid and diverse planning trajectories may exist for a single task.

To counteract this limitation, we propose Trust Region Stretch (TRS), which relaxes the clipping bounds in PPO’s surrogate objective, thereby expanding the allowable update range. This encourages higher variance in trajectory exploration and mitigates the tendency toward mode collapse. The modified objective becomes:

\begin{equation}
L^{\text{TRS}}(\theta) = \mathbb{E}_{t} \left[ 
\min \left( 
r_t(\theta) \cdot \hat{A}_t,\ 
\text{clip}(r_t(\theta),\ \epsilon_{\text{low}},\ \epsilon_{\text{high}}) \cdot \hat{A}_t 
\right) 
\right]
\end{equation}

where \( \hat{A}_t \) is the estimated advantage at timestep \( t \), and \( \epsilon_{\text{low}}, \epsilon_{\text{high}} \) denote the relaxed clipping bounds (e.g., \( \epsilon_{\text{low}}=0.6 \), \( \epsilon_{\text{high}}=1.8 \)).

\paragraph{(2) Precision Token Loss.}  
The original GRPO~\cite{shao2024deepseekmath} framework computes loss at the sample level by first averaging token-level losses within each sequence, followed by aggregation across samples. In CAD practice, different tokens play markedly different roles in shaping geometric accuracy. Drawing errors such as numerical dimensions have disproportionately severe impacts on the correctness of the resulting 3D model. 
To address this, we introduce Precision Token Loss, a weighted loss function that amplifies gradients on semantically important tokens. Let \( \omega_t \) denote the importance weight for token \( s_t \), determined based on token type or positional heuristics. The loss over a batch \( \mathcal{B} \) is computed as:

\begin{equation}
\mathcal{L}_{\text{precision}} = \frac{1}{|\mathcal{B}|} \sum_{i=1}^{|\mathcal{B}|} \left( \frac{1}{Z^{(i)}} \sum_{t=1}^{|s^{(i)}|} \omega_t \cdot \ell(s^{(i)}_t) \right)
\end{equation}

where \( \ell(s^{(i)}_t) \) is the token-level negative log-likelihood, and \( Z^{(i)} = \sum_{t=1}^{|s^{(i)}|} \omega_t \) is a normalization factor to ensure that the per-sample loss magnitude remains consistent. For tokens corresponding to numerical parameters or geometry-affecting operations, we set \( \omega_t > 1 \); otherwise, \( \omega_t = 1 \).

\paragraph{(3) Overlong Filtering.}  
In reinforcement learning for autoregressive sequence generation, it is standard practice to set a maximum generation length \( T_{\text{max}} \) to avoid unbounded outputs. Sequences exceeding this limit are typically truncated and penalized, either by assigning zero reward or by applying harsh penalties. However, such treatment can introduce substantial noise into the reward signal, especially in tasks like CAD modeling where output verbosity does not always imply functional error.
To mitigate this issue, we propose an Overlong Filtering strategy that excludes truncated samples from reward computation entirely. Formally, let \( \mathcal{S}_{\text{trunc}}\subset \mathcal{S} \) denote the set of truncated sequences in a training batch. We modify the reinforcement learning objective to:

\begin{equation}
\mathcal{L}_{\text{RL}} = -\mathbb{E}_{s \sim \mathcal{S} \setminus \mathcal{S}_{\text{trunc}}} \left[ R(s) \cdot \log \pi_\theta(s) \right]
\end{equation}

\subsection{ExeCAD}
\label{3.4}
A key obstacle hindering progress in CAD code generation is the absence of high-quality, open-source datasets that accurately reflect the precision and constraints of industrial design. Existing datasets frequently exhibit semantic misalignment between natural language descriptions and CAD code, resulting in inconsistencies such as vague or incomplete specifications, incorrect geometric references, and incoherent parametric logic. These issues limit the effectiveness of supervision and degrade the model’s ability to align symbolic code with geometric semantics.

To overcome these limitations, we introduce \textbf{ExeCAD}—a large-scale, semantically aligned dataset specifically curated for executable and editable CAD modeling tasks. ExeCAD comprises 16,540 high-fidelity samples, each consisting of:
\begin{itemize}
    \item A natural language prompt simulating non-expert design input;
    \item A structured expert-level design language specification;
    \item Executable and editable CadQuery code;
    \item A rendered 3D CAD model as ground-truth geometry.
\end{itemize}

At the core of ExeCAD’s construction is a bidirectional semantic refinement pipeline designed to ensure tight consistency between modalities. This pipeline aligns geometric structure, parametric meaning, and design logic by iteratively correcting and reconstructing both the textual and code modalities.

\begin{table*}[t]
\centering
\small
\setlength{\tabcolsep}{3.5pt} 
\caption{Comparison on the ExeCAD benchmark using two types of input text: Natural Language Description and Structured Design Language. The upper block reports results using pretrained leading VLMs without fine-tuning; the lower block shows results after fine-tuning. ``Med CD'' reflects the median of CD. CD metrics are $\times 10^3$. IOU and Executability are higher-is-better; CD metrics are lower-is-better. ``Reference'' indicates the reference image; ``Exec.'' indicates the executability. $\ast$ indicates our re-implementation trained on the same benchmark.}
\label{result}
\begin{tabular}{l|cccc|cccc}
\toprule
 \multirow{2}{*}{\textbf{\textsc{Model}}} & \multicolumn{4}{c|}{\textbf{Natural Language Description}}  & \multicolumn{4}{c}{\textbf{Structured Design Language}} \\
& IOU(\%)$\uparrow$ & Mean CD$\downarrow$  & Med CD$\downarrow$ &Exec.(\%)$\uparrow$  & IOU(\%)$\uparrow$ & Mean CD$\downarrow$  & Med CD$\downarrow$ &Exec.(\%)$\uparrow$\\

  \midrule
  LLaVA-1.5~\cite{liu2024improved} & 0.2  & 178.19 & 110.76 & 1.67 & 0.0  & 144.76 & 91.30 & 0.13 \\
Phi-3.5-Vision~\cite{abdin2024phi} & 4.4  & 192.09  & 121.05  & 32.50  & 6.7 & 161.9 & 93.45 & 51.63 \\
InternVL2.5~\cite{chen2024internvl} & 13.7 & 161.75 & 86.56 & 55.17 & 19.5 & 136.29 & 72.13 & 72.33 \\
Qwen2.5-VL~\cite{qwen2.5-VL} & 30.6 & 103.70 & 1.95 & 52.47 & 31.6 & 114.64 & 7.70 & 62.30 \\
Claude4~\cite{anthropic2024claude35}& 38.9 & 92.81 & 4.96 & 63.26 & 41.8 & 91.29 & 4.37 & 64.42 \\
InternVL3~\cite{zhu2025internvl3exploringadvancedtraining}& 40.9 & 82.59 & 2.57 & 56.72 & 43.5 & 82.72 & 2.15 & 72.03   \\
Gemini2.5 Pro~\cite{gemini2.5pro2024}& 41.8 & 83.23 & 2.43 & 64.39 & 45.1 & 79.93 & 2.06 & 66.13  \\
GPT-4o~\cite{achiam2023gpt} & 47.3 & 77.28 & 1.76 & 73.16& 47.7 & 65.52 & 1.47 & 72.72 \\
\midrule
CAD-Coder~\cite{guan2025cad} $\ast$ & 66.2 & 41.96 & 1.17 & 94.72 & 70.4 & 29.63 & 1.02 & 95.44  \\
Ours & \textbf{72.3} & \textbf{33.68} & \textbf{0.68} &  \textbf{99.63}&  \textbf{78.7} & \textbf{15.21} & \textbf{0.63} & \textbf{98.83} \\
\bottomrule
\end{tabular}
\end{table*}

\paragraph{Dataset Construction Pipeline.}
We construct ExeCAD by refining two open-source datasets that are independently derived from the DeepCAD project: GenCAD-Code~\cite{alam2024gencad}, which provides syntactically valid but dimensionally noisy CadQuery scripts, and Text2CAD~\cite{khan2024text2cad}, which contains human-written natural language descriptions that are often vague, redundant, or semantically inconsistent.
To ensure high-quality multimodal supervision, we employ GPT-4o as an alignment assistant and guide it through four refinement stages using prompt engineering:

\begin{itemize}
    \item \textbf{Input Preparation:} We initialize the pipeline by pairing noisy CADQuery code from GenCAD-Code with unstructured or loosely aligned natural language prompts from Text2CAD.
    
    \item \textbf{Text Normalization:} The natural language descriptions are cleaned to remove redundancy and vagueness, producing refined general-purpose natural language descriptions that are syntactically simple yet semantically complete for non-expert use.
    
    \item \textbf{Code Correction:} Based on the normalized text, the corresponding CadQuery code is adjusted to align with geometric intent. This includes correcting dimensional parameters, feature ordering, and topological constraints—ensuring strict geometric-semantic alignment.
    
    \item \textbf{Semantic Reconstruction:} To guarantee full coverage of geometric and parametric semantics, we generate a reverse-mapped, expert-oriented structured design language from the corrected code. This textual variant explicitly encodes geometry primitives, construction sequences, constraint logic, and numeric parameter intent.
\end{itemize}

The resulting dataset ensures structural and semantic coherence across all modalities—natural language, design logic, CAD code, and rendered geometry. This makes ExeCAD a robust benchmark for training and evaluating multimodal CAD code generation systems under varying input formats and user expertise levels.

\section{Experiments}

\subsection{Implementation Details}
All experiments are conducted on NVIDIA A100 GPUs. Most experiments use Qwen2.5-VL as the base model, trained on a single server equipped with 8 A100 GPUs and a batch size of 64. We use the AdamW optimizer with a cosine annealing scheduler. The learning rate are 1e-6 for RL post training and 2e-5 for CoT-based Cold Start.

\begin{table*}[t]
\centering
\small
\setlength{\tabcolsep}{4pt} 
\caption{Ablation study of CAD-RL on the ExeCAD benchmark.  
We investigate the impact of incorporating Reference Images, Chain-of-Thought (CoT) reasoning, and Reinforcement Learning (RL) during training.  
Results are reported separately under two input settings: Natural Language Description and Structured Design Language. CD metrics are $\times 10^3$.
$\uparrow$ indicates higher is better; $\downarrow$ indicates lower is better; ``Ref img'' indicates the reference image; ``Exec.'' indicates the executability.
}
\label{aba}
\begin{tabular}{cccc|cccc|cccc}
\toprule
 \multirow{2}{*}{Cold Start} & \multirow{2}{*}{Ref img} & \multirow{2}{*}{CoT}& \multirow{2}{*}{RL} & \multicolumn{4}{c|}{\textbf{Natural Language Description}}  & \multicolumn{4}{c}{\textbf{Structured Design Language}} \\
&&&& IOU(\%)$\uparrow$ & Mean CD$\downarrow$  & Med CD$\downarrow$ &Exec.(\%)$\uparrow$  & IOU(\%)$\uparrow$ & Mean CD$\downarrow$  & Med CD$\downarrow$ &Exec.(\%)$\uparrow$\\
\midrule
\checkmark & & &  & 49.5 &74.27 &1.67 &90.64 &50.3 &66.73 &1.62 &91.86       \\

\checkmark& \checkmark &&&50.1 &72.96 &1.74 &90.25 &51.7 & 66.48 & 1.61 & 91.61 \\

\checkmark & &\checkmark &&60.8 &66.09 &1.48 &94.41 &63.3 & 59.01 & 1.34 & 94.89\\

\checkmark & \checkmark &\checkmark &&63.9 &63.36 &1.36 &94.66 &64.5 & 44.70 & 1.21 & 94.84\\
\checkmark & \checkmark &\checkmark &\checkmark&\textbf{72.3} &\textbf{33.68} &\textbf{0.68} &\textbf{99.63} &\textbf{78.7} & \textbf{15.21} & \textbf{0.63} & \textbf{98.83}\\
\bottomrule
\end{tabular}
\end{table*}

\subsection{Datasets and Metrics}
We evaluated both the leading vision–language models (VLMs) and our proposed method on ExeCAD.
We adopt four complementary metrics to evaluate the quality of CAD code generation, focusing on geometric accuracy, code executability, and structural precision:
(1) Intersection-over-Union (IoU): Evaluates the volumetric overlap between predicted and reference shapes. Due to its strict requirement for spatial alignment, IoU is particularly suitable for assessing shape accuracy in industrial CAD scenarios.
(2) Mean Chamfer Distance (Mean CD): Measures the average geometric difference between the generated and ground-truth models in point cloud space. It reflects the overall shape fidelity.
(3) Median Chamfer Distance (Median CD): Captures the typical per-sample error while being more robust to outliers, providing a stable estimate of model performance.
(4)Executability: This directly reflects the functional validity of the output.

\subsection{Benchmarking CAD Code Generation}

Table~\ref{result} presents a comprehensive evaluation on the ExeCAD benchmark across two types of textual inputs: Natural Language Description and Structured Design Language. The comparison includes eight state-of-the-art pretrained vision-language models (VLMs)—spanning both open-source models such as Qwen2.5-VL~\cite{qwen2.5-VL} and InternVL3~\cite{zhu2025internvl3exploringadvancedtraining}, as well as proprietary models like GPT-4o~\cite{achiam2023gpt}—all evaluated without any task-specific post training. We also benchmark two fine-tuned baselines: CAD-Coder~\cite{guan2025cad} and our proposed CAD-RL.

Overall, CAD-RL achieves the best performance across all metrics. It reaches a code executability of 99.63\% and an IoU of 78.7\% under Structured Design Language input, surpassing CAD-Coder by +8.3\% in IoU and +3.39\% in executability, demonstrating the effectiveness of our post training method.

\noindent\textbf{1) Pretrained VLMs face significant challenges in CAD code generation.}
While VLMs show strong capabilities on general multimodal tasks, Table~\ref{result} reveals their poor performance on CAD-specific code generation. Even the best-performing pretrained model (GPT-4o) only achieves a 72.72\% executability and 47.7\% IoU. This limitation stems from the inherent complexity of CAD modeling, which demands both symbolic reasoning over operations (e.g., extrusion, chamfer) and precise understanding of numeric constraints. These capabilities are far beyond the general representation learned by existing VLMs, necessitating task-specific adaptation.

\noindent\textbf{2) Structured design input significantly improves generation quality.}
Across both pretrained and fine-tuned models, Structured Design Language yields consistently higher performance than Natural Language Descriptions. This is because structured inputs are less ambiguous and contain less redundancy, enabling more precise alignment with the underlying geometry and code structure. Nevertheless, real-world usage often involves mixed or informal text input—highlighting the importance of supporting both modalities.

\noindent\textbf{3) Size precision remains a key challenge.}
CAD-RL achieves 98.83\% executability under structured input, indicating that the model has fully learned to generate syntactically and semantically valid CADQuery code. However, its IoU remains at 78.7\%, exposing residual imprecision in the spatial structure. This performance gap underscores that while the model grasps the correct parametric form, fine-grained control over geometric dimensions remains challenging. This observation further validates our introduction of Precision Token Loss, which explicitly targets the numerical sensitivity of such tokens.

\noindent\textbf{4) Metric selection matters: Mean CD vs. Median CD vs. IoU.}
Following CAD-Coder~\cite{guan2025cad}, we adopt both the mean and median Chamfer Distance (CD) to evaluate shape similarity. While Mean CD is widely used, it is highly sensitive to outliers—especially when large parts are under-sampled due to limited point resolution, leading to disproportionate distortion of the average. Median CD, on the other hand, captures typical cases and offers a more robust estimation. However, it fails to reflect subtle but critical structural inaccuracies. For instance, Claude 4~\cite{anthropic2024claude35} reports a better IoU than Qwen2.5-VL~\cite{qwen2.5-VL} under natural language input, yet its median CD is significantly worse.
To address this discrepancy, we incorporate \textbf{IoU} as an additional metric. Unlike perceptual or distributional distances, IoU directly enforces near-exact volumetric overlap, aligning more closely with industrial design requirements where precise geometric conformity is critical. Given the low tolerance for structural drift in CAD applications, IoU provides a more faithful measure of functional shape accuracy.

\subsection{Ablation Study}
To better understand the contribution of each component in CAD-RL, we conduct a systematic ablation study on the ExeCAD benchmark. As shown in Tab.~\ref{aba}, we progressively incorporate three key parts: Reference Images, Chain-of-Thought (CoT), and Reinforcement Learning (RL) Post Training. Results are reported under both Natural Language Description and Structured Design Language input settings.

When using only cold-start supervised fine-tuning (SFT) without any reasoning or visual guidance, the model already achieves high code executability—90.64\% and 91.86\% under natural and structured text, respectively. However, the IOU remains low (around 50\%), indicating that while the model learns to produce syntactically valid CADQuery code, it fails to accurately capture geometric dimensions and constraints—resulting in imprecise shape reconstructions.

Incorporating Chain-of-Thought reasoning substantially boosts IOU and geometry fidelity. For instance, under natural language inputs, IOU improves from 49.5\% to 60.8\%. This suggests that multi-step planning enables the model to reason through construction logic more effectively.

Adding reference images alone brings modest gains, but when used alongside CoT, performance improves notably (e.g., IOU rises from 60.8\% to 63.9\%). We hypothesize that visual input helps resolve ambiguities or omissions in the textual description, especially when the model already possesses strong planning capabilities via CoT.

Among all components, RL leads to the most substantial improvement. Under natural language input, RL raises IOU from 63.9\% to 72.3\%, with Mean CD dropping to 33.68 and Med CD reaching a minimum of 0.68—establishing new state-of-the-art results. This confirms that reward-driven optimization effectively balances executability and accuracy, pushing the model toward structurally faithful and geometrically precise CAD programs.

By integrating CoT, visual grounding, and RL-based optimization, CAD-RL achieves strong reasoning, perception, and decision-making capabilities—culminating in superior CAD code generation performance across both geometric and functional metrics.

\section{Conclusion}
In this paper, we presented CAD-RL, a reinforcement learning framework for precise and executable CAD code generation from multimodal inputs. By leveraging text description and reference image, our method integrates Chain-of-Thought reasoning with reward-driven post training to align human design intent with syntactically and semantically correct CAD programs.
To support evaluation and training, we introduced ExeCAD, a large-scale benchmark of 16,540 high-quality triplets encompassing text, code, and geometry. Experimental results demonstrate that CAD-RL achieves SOTA performance in both code executability and geometric fidelity, outperforming leading open-source and proprietary baselines. Our ablation studies further confirm the synergistic contributions of visual grounding, step-wise reasoning, and fine-grained reward functions, such as IoU-based alignment and precision-aware penalties.
This work underscores the feasibility and promise of multimodal CAD automation workflow and highlights the critical role of reasoning-guided learning in bridging abstract design instructions with formal CAD representations. Future research may extend this framework to more expressive 3D domains, incorporate real-world manufacturing constraints, and explore interactive design paradigms.

\bigskip

\bibliography{aaai2026}

\end{document}